# ROFT: Real-time Optical Flow-aided 6D Object Pose and Velocity Tracking

Nicola A. Piga[1,2], Yuriy Onyshchuk[1], Giulia Pasquale[1], Ugo Pattacini[1], and Lorenzo Natale[1]

*Abstract*—6D object pose tracking has been extensively studied in the robotics and computer vision communities. The most promising solutions, leveraging on deep neural networks and/or filtering and optimization, exhibit notable performance on standard benchmarks. However, to our best knowledge, these have not been tested thoroughly against fast object motions. Tracking performance in this scenario degrades significantly, especially for methods that do not achieve real-time performance and introduce non negligible delays. In this work, we introduce ROFT, a Kalman filtering approach for 6D object pose and velocity tracking from a stream of RGB-D images. By leveraging real-time optical flow, ROFT synchronizes delayed outputs of low frame rate Convolutional Neural Networks for instance segmentation and 6D object pose estimation with the RGB-D input stream to achieve fast and precise 6D object pose and velocity tracking. We test our method on a newly introduced photorealistic dataset, Fast-YCB, which comprises fast moving objects from the YCB model set, and on the dataset for object and hand pose estimation HO-3D. Results demonstrate that our approach outperforms state-of-the-art methods for 6D object pose tracking, while also providing 6D object velocity tracking. A video showing the experiments is provided as supplementary material.

## I. INTRODUCTION

THE ability to visually identify objects in the scene and estimate their pose is fundamental for autonomous robotic systems. Several approaches have been proposed to tackle the problems of 6D object pose estimation [1], [2], [3], refinement [4], [5] and tracking [6], [7].

Although object pose estimation might suffice to initiate an interaction with the environment, the ability to track the pose and the velocity of the object over consecutive frames is of paramount importance for purposeful object manipulation [8].

Recent works on 6D object pose tracking, based on deep learning [7] or on its combination with Kalman and particle filtering [6], showed remarkable performance on standard datasets such as YCB-Video [1]. Nevertheless, these datasets comprise scenes with slowly moving object poses that might be inadequate for benchmarking 6D object pose tracking algorithms. Furthermore, most methods overlook the fact that, in case of limited computational resources or due to architectural complexity, large computation times induce non-negligible delays and sparsity in the output, which negatively affect the overall tracking process. Additionally, dedicating large amount of computing solely to perform object tracking may be impractical because this task is typically only one among the several that have to be executed simultaneously on a robot.

In this paper, we propose to combine real-time optical flow with *low frame rate* Convolutional Neural Networks (CNNs) for instance segmentation and 6D object pose estimation to achieve real-time 6D object pose and velocity tracking from RGB-D images. Our contributions are the following:

- We show how to exploit real-time optical flow to synchronize *delayed* instance segmentation and 6D object pose estimation streams with a given RGB-D input stream.
- We combine optical flow and the aforementioned synchronized streams in a Kalman filtering approach that tracks both the 6D pose and the 6D velocity of a given object.
- We propose a synthetic photorealistic dataset, Fast-YCB, comprising scenes with fast moving objects from the YCB model set [9] with linear velocities up to 63 cm/s and rotational velocities up to 266 degrees/s.

We validate our method experimentally on two datasets: the newly introduced Fast-YCB dataset and the publicly available dataset for object and hand pose estimation HO-3D [10]. Experiments show that our method outperforms the output of the 6D pose estimation network alone and, more importantly, state-of-the-art methods for object pose tracking.

The Fast-YCB dataset will be published together with the software implementation to ensure reproducibility.

The rest of the paper is organized as follows. After a section on related work, we present our pipeline, followed by a section on experimental evaluations and a conclusion.

## II. RELATED WORK

Recent works addressing the 6D object pose tracking problem from RGB-D images rely on deep neural networks. se(3)-TrackNet [7] is an end-to-end network which provides the relative pose given the current RGB-D observation and a synthetic rendering of the previous estimate. Tracking is performed by propagating over time a given initial condition using the predicted relative poses. PoseRBPF [6] combines particle filtering with a deep autoencoder in order to estimate the full object pose distribution. These methods either do not model the object motion using a specific motion prior or adopt a simple constant velocity model that might be inadequate in case of fast object motions.

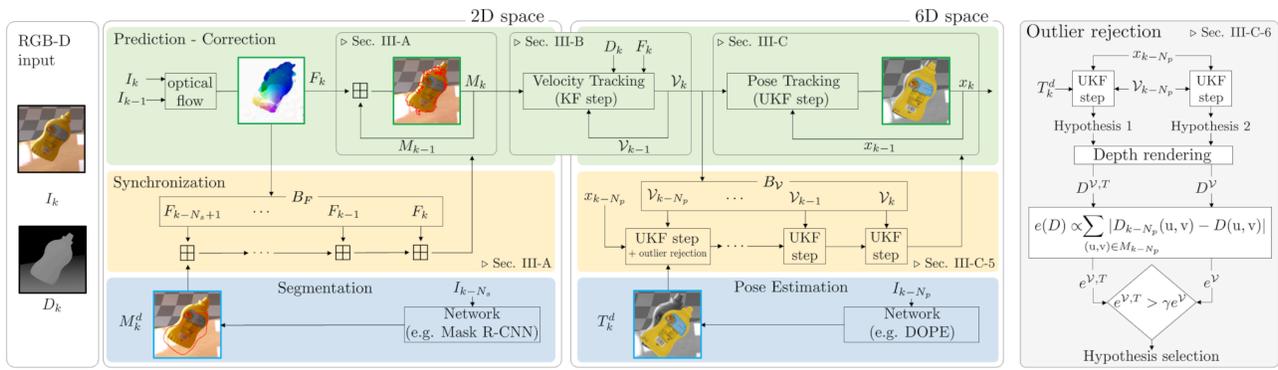

Fig. 1: Overview of our pipeline for 6D object pose tracking.

In this respect, other methods have been proposed that integrate information on actual pixels motion, obtained from optical flow, within 6D object pose tracking. Early works that use contour-based 2D-3D correspondences to track the 6D object pose over time [11], [12] propose to use optical flow to improve contour matching from current to the next frame even in presence of large object displacements.

Considering the integration between deep neural networks and optical flow, DeepIM [13] leverages the FlowNet [14] network for optical flow prediction to iteratively refine 6D object poses, similarly to [7]. Following a different path, Zhang et al. [15] combine recent advances in CNNs for instance segmentation, optical flow and depth information to track ego and object 6-DoF motion and linear velocity. However, their work mostly focuses on automotive scenarios that are quite different from those considered in this work.

In our work, we combine CNNs with filtering algorithms, similarly to PoseRBPF [6]. However, we explicitly take into account information from optical flow to alleviate the weaknesses of simple constant velocity motion models. Inspired by [15], we combine instance segmentation and optical flow for 6D object pose tracking. However, we explicitly account for delayed and low frame rate instance segmentation in case of limited computational resources, and we propose an optical flow-based synchronization mechanism. Differently from all the mentioned works, our approach provides an explicit estimate of the linear and *angular* velocities of the object, that have been shown to be valuable for closed-loop control and object manipulation [8].

## III. METHODOLOGY

Given an input stream of RGB $\{I_k\}$ and depth $\{D_k\}$ images and an initial estimate of the object pose, the aim of our pipeline is to track the 6D pose and the 6D velocity of the object with respect to the camera frame for every frame $k$. We compute a stream of optical flow frames $\{F_k\}$ from consecutive RGB frames $I_{k-1}, I_k$, and assume that the following are available: i) a stream of segmentation masks $\{M_k\}$ of the object of interest, ii) a stream of pose measurements $\{T_k \in SE(3)\}$.

$M_k$ and $T_k$ can be obtained using either separate deep learning-based networks for segmentation and 6D object pose estimation, or a single network which performs both tasks jointly [1], [2].

We are interested in the scenario where the network outputs are only available at *low frame rate* resulting in non-synchronized and delayed streams $\{M_k^d\}$ and $\{T_k^d\}$.

Our pipeline consists of three stages that we detail in the sections III-A, III-B and III-C. An overview of our pipeline is depicted in Fig. 1.

### A. Optical flow-aided instance segmentation

Given an input RGB stream $\{I_k\}$, we exploit the optical flow frames $\{F_k\}$ synchronized with $\{I_k\}$ and a low frame rate and non-synchronized stream of segmentation masks $\{M_k^d\}$ to produce a stream of masks $\{M_k\}$ that is synchronized with $\{I_k\}$.

Let $M \boxplus F$ be the operator

$$M \boxplus F = \{(u,v) + F(u,v) \quad \forall (u,v) \in M\}, \quad (1)$$

which propagates the pixels coordinates in the mask $M$ according to the flow vectors $F(u,v) \in \mathbb{R}^2$.

We initialize the stream $\{M_k\}$ using the the first available mask $M_0^d$, under the assumption that the object is stationary at time $k = 0$. In order to provide a continuous and synchronized stream of masks $\{M_k\}$, we propagate it over time using the optical flow

$$M_k = M_{k-1} \boxplus F_k, \quad (2)$$

until a new mask $M^d$ from the network is available. Meanwhile we store the frames $F_k$ in a buffer $B_F$. After $N_s$ steps, we assume that a new mask $M_k^d$, associated to the RGB frame $I_{k-N_s}$, is available. Since the received mask is associated to a *past* RGB frame, we iteratively propagate it forward using the buffered flow frames from $B_F$ up to time $k$

$$M_k = \left(\left(\left(M_k^d \boxplus F_{k-N_s+1}\right) \boxplus \ldots\right) \boxplus F_{k-1}\right) \boxplus F_k, \quad (3)$$

thus obtaining a synchronized mask $M_k$. From here, we proceed again with the propagation scheme in Eq. (2), until a new mask from the network is available again. The overall process is illustrated in the "Synchronization" block in Fig. 1.

We remark that the synchronized masks might be discontinuous due to the noise in the optical flow vectors. Nevertheless, our approach relies more on the availability of pixel coordinates in the image $I_k$ belonging to the surface of the object rather than the exact shape of the segmentation mask.

### B. 6D object velocity tracking

Having at disposal a stream of depth images $\{D_k\}$, optical flow frames $\{F_k\}$ and synchronized segmentation masks $\{M_k\}$, we track the 6D object velocity $\mathcal{V}$ from optical flow measurements using a linear Kalman Filter.

*1) State definition:* We draw from the Screw Theory [16] and we define the state $\mathcal{V}$ as

$$\mathcal{V} = \begin{bmatrix} v_O & \omega \end{bmatrix} \quad (4)$$

where $\omega \in \mathbb{R}^3$ is the angular velocity of the object and $v_O \in \mathbb{R}^3$ is the velocity of a point which coincides with the origin of the camera instantaneously and moves as it was rigidly attached to the object.

*2) Motion model:* We assume that the underlying dynamics of the state vector $\mathcal{V}$ is described by the simple motion model

$$\begin{aligned} \mathcal{V}_k &= \mathcal{V}_{k-1} + w, \\ w &\sim \mathcal{N}(0, \mathrm{diag}(Q_v, Q_\omega)) \end{aligned} \quad (5)$$

where velocity increments $\mathcal{V}_k - \mathcal{V}_{k-1}$ are Gaussian with covariances $Q_v \in \mathbb{R}^{3\times 3}$ and $Q_\omega \in \mathbb{R}^{3\times 3}$ associated to the linear and angular velocity respectively.

*3) Measurement model:* We define the optical flow measurements as the vector

$$y_k^F(F_k, M_{k-1}) = \begin{bmatrix} \dots \\ F_k(\mathrm{u},\mathrm{v}) \\ \dots \end{bmatrix} \quad (\mathrm{u},\mathrm{v}) \in M_{k-1} \quad (6)$$

collecting the flow vectors $F_k(\mathrm{u},\mathrm{v}) \in \mathbb{R}^2$ for all the pixels belonging to the segmentation mask. We use the mask at time $k-1$, instead of time $k$, because the optical flow vector field describes the motion of points from frame $k-1$ to frame $k$. In order to relate the measurements in Eq. (6) to the state $\mathcal{V}_k$, we build upon the standard pinhole camera model

$$\begin{bmatrix} \mathrm{u} \\ \mathrm{v} \end{bmatrix} = \begin{bmatrix} c_x + \frac{x}{d} f_x \\ c_y + \frac{y}{d} f_y \end{bmatrix} \quad (7)$$

that relates a pixel $(\mathrm{u},\mathrm{v})$ in the RGB image with its 3D counterpart $(x, y, d)$ where $d \in \mathbb{R}$ is the depth associated to the pixel $(\mathrm{u},\mathrm{v})$ extracted from the stream $\{D_k\}$. Differentiating both sides of Eq. (7) with respect to time and assuming the rigidity of the object, it can be shown that the measurements $y_k^F$ can be predicted as

$$\begin{aligned} y_k^F &= h^F(\mathcal{V}_k, M_{k-1}) = J(M_{k-1})\mathcal{V}_k + \nu, \\ \nu &\sim \mathcal{N}(0, R_F) \end{aligned} \quad (8)$$

where $\nu$ represents additive Gaussian noise with measurement noise covariance $R_F$. The jacobian $J(M) \in \mathbb{R}^{2|M|\times 6}$ maps the 6D velocity of the object to the flow vectors for all the pixels in the mask $M$ and is given by

$$J(M) = \begin{bmatrix} \dots \\ J_{\mathrm{u},\mathrm{v}} \\ \dots \end{bmatrix} \quad (\mathrm{u},\mathrm{v}) \in M, \quad (9)$$

$$J_{\mathrm{u},\mathrm{v}} = \begin{bmatrix} J_{v_O} & J_\omega \end{bmatrix} \Delta_T, \quad J_{v_O} = \begin{bmatrix} \frac{f_x}{d} & 0 & -\frac{(\mathrm{u}-c_x)}{d} \\ 0 & \frac{f_y}{d} & -\frac{(\mathrm{v}-c_y)}{d} \end{bmatrix},$$

$$J_\omega = \begin{bmatrix} \frac{-(\mathrm{u}-c_x)(\mathrm{v}-c_y)}{f_y} & \frac{f_x^2+(\mathrm{u}-c_x)^2}{f_x} & \frac{-(\mathrm{v}-c_y)f_x}{f_y} \\ \frac{-f_y^2-(\mathrm{v}-c_y)^2}{f_y} & \frac{(\mathrm{u}-c_x)(\mathrm{v}-c_y)}{f_x} & \frac{(\mathrm{u}-c_x)f_y}{f_x} \end{bmatrix}$$

where $\Delta_T$ is the time elapsed between consecutive frames.

*4) Kalman filtering:* Models in Eqs. (5) and (8) are linear in the state $\mathcal{V}_k$, hence they can be employed within a linear Kalman Filter [17] in order to track the object velocity $\mathcal{V}_k$. The state update equations are not reported for brevity and can be found in [17].

### C. 6D object pose tracking

In this section we present the last stage of our pipeline which fuses non-synchronized pose measurements $\{T_k^d\}$, available at low frame rate, with the estimated velocities $\{\mathcal{V}_k\}$, available for all frames, using an Unscented Kalman Filter in order to track the 6D pose of the object. In the following subsections we detail the filtering process.

*1) State definition:* We define the state $x$ to be tracked over time as

$$x = [t, v, q, \omega] \quad (10)$$

which comprises the Cartesian position $t \in \mathbb{R}^3$, a unitary quaternion $q$ for the 3D orientation, the linear velocity $v \in \mathbb{R}^3$, not to be confused with the pixel coordinate v, and the angular velocity $\omega \in \mathbb{R}^3$.

*2) Motion model:* We assume that the state $x_k$ evolves according to the following constant velocity model

$$\begin{bmatrix} t_k \\ v_k \\ q_k \\ \omega_k \end{bmatrix} = \begin{bmatrix} t_{k-1} + v_{k-1}\Delta_T \\ v_{k-1} \\ A_q(\omega_{k-1})q_{k-1} \\ \omega_{k-1} \end{bmatrix} + w, \quad (11)$$

$$w \sim \mathcal{N}(0, \mathrm{diag}(Q_t, 0, Q_q))$$

where $A_q(\omega)$ is the standard quaternion kinematics transition matrix [18] and $Q_t \in \mathbb{R}^{6\times 6}$, $Q_q \in \mathbb{R}^{3\times 3}$ are the noise covariance matrix for the translational and rotational components of the state respectively. Following prior work [18], the process noise affects the quaternion $q$ only indirectly via the angular velocity $\omega$.

*3) Measurement model:* We collect pose and velocity measurements in a vector

$$y_k(T_k^d, \mathcal{V}_k) = \begin{bmatrix} p(T_k^d) \\ q(T_k^d) \\ v_O(\mathcal{V}_k) \\ \omega(\mathcal{V}_k) \end{bmatrix} \quad (12)$$

where $p(T_k^d) \in \mathbb{R}^3$, $q(T_k^d)$ are the Cartesian position and the quaternion components of the measured pose $T_k^d$ and $v_O(\mathcal{V}_k) \in \mathbb{R}^3$, $\omega(\mathcal{V}_k) \in \mathbb{R}^3$ are the linear and angular velocity components of the measured velocity $\mathcal{V}_k$. The measurement model relating $y_k$ with the state vector $x_k$ is expressed as

$$y_k(x_k) = \begin{bmatrix} t_k \\ q_k \\ v_k + t_k \times \omega_k \\ \omega_k \end{bmatrix} \oplus \nu, \quad (13)$$

$$\nu \sim \mathcal{N}(0, \mathrm{diag}(R_t, R_q, R_v, R_\omega))$$

with $R_* \in \mathbb{R}^{3\times 3}$ the noise covariance matrix associated to each component and $\oplus$ representing the operation of adding noise taking into account the quaternion arithmetic [18]. We remark that if pose measurements are not available at time $k$, the equation simplifies to the last two rows resulting in a measure of the object velocity only.

*4) Unscented Kalman filtering:* Actual tracking of the state $x_k$ is performed using a quaternion-based Unscented Kalman Filter [18]. This choice is dictated by the non-linearity of the models in Eqs. (11), (13) and the adoption of quaternions in both the state and the measurements.

At each step $k$, the previous state is propagated through the motion model in Eq. (11) to obtain the *predicted* state $x_k^-$. Then, a correction step incorporates the measurements $y_k$ in the state belief $x_k$ according to the measurement model in Eq. (13). The UKF state update equations are not reported for brevity and can be found in [18], [19].

*5) Synchronization of pose measurements:* Pose measurements $\{T_k^d\}$ are available at low frame rate. We assume that a new pose is available each every $N_p$ steps, hence it is associated to the RGB frame $I_{k-N_p}$ and could not be used directly as measurement at time $k$. For this reason, when a new pose measurement $T_k^d$ becomes available, we reset the filter to the previous state $x_{k-N_p}$ and we perform a UKF update using $T_k^d$ and the velocity $\mathcal{V}_{k-N_p}$ as measurements. We assume that previous velocity measurements are stored in a buffer $B_\mathcal{V}$. We then perform additional UKF updates using the buffered velocities from $B_\mathcal{V}$ up to time $k$ in order to obtain a synchronized estimate $x_k$. The process is illustrated in the "Synchronization" block in Fig. 1.

*6) Rejection of pose measurements outliers:* Pose measurements $\{T_k^d\}$ from CNNs are often affected by outliers which might induce biases in the estimates $x_k$. To overcome this issue, when using $T_k^d$ as a measurement, we actually perform two UKF updates, one considering both pose and velocity measurements, the other considering only the velocity. Assuming that a 3D mesh of the object is available, we then render two synthetic depth maps, $D^{\mathcal{V},T}$ and $D^\mathcal{V}$, for both estimates, and compare to the measured depth $D_{k-N_p}$. If the resulting errors $e(D)$ differ considerably, according to a threshold $\gamma$, the pose $T_k^d$ is marked as outlier and skipped. For additional details and the definition of $e(D)$ we refer to the "Outlier rejection" block in Fig. 1.

## IV. EXPERIMENTAL SETUP

Our pipeline relies on segmentation and pose estimation as inputs. As source of segmentation masks $\{M_k^d\}$, we adopt the state-of-the-art Mask R-CNN instance segmentation network [20] pre-trained on the COCO dataset [21]. In order to obtain a segmentation model for the objects of interest from the YCB model set [9], we fine-tuned the network on a synthetic dataset which was generated using the BOP BlenderProc renderer [22]. Object poses $\{T_k^d\}$ are obtained using the 6D object pose estimation network DOPE, originally trained on purely synthetic data, which has shown the ability to generalize to novel environments [23].

In our experiments we compare our pipeline against two recent deep learning-based methods for 6D object pose tracking from RGB-D images, recognized as the state of the art: PoseRBPF [6] and se(3)-TrackNet [7].

Since all the aforementioned methods were originally trained on purely synthetic images comprising the objects of interest, we do not retrain DOPE, PoseRBPF and se(3)-TrackNet and use the original weights provided by the authors.

### A. Evaluation datasets

Several standard datasets for 6D object pose estimation have been proposed in the literature such as T-LESS [24], LineMOD [25] and YCB-Video [1]. Among them, YCB-Video has also been adopted to benchmark 6D object pose trackers such as PoseRBPF and se(3)-TrackNet. However, the provided sequences are characterized by slowly varying poses and low velocities, hence not ideal to benchmark 6D object pose tracking algorithms.

**Fast-YCB.** For these reasons, we created a new dataset[1], that we called Fast-YCB, consisting of six photorealistic synthetic sequences, each comprising fast motions of a single object taken from the YCB model set [9] in a tabletop scenario. The six objects have been selected in order to consider a variety of shapes (box, cylinders or irregular), sizes (small, medium and large) and textures (see Fig. 2). Each sequence is rendered with bright static light conditions and provided with 30 frames per second (fps) 1280x720 RGB-D frames and exact ground-truth 6D object poses and velocities. Optical flow frames are also provided as part of the dataset. We remark that object trajectories, used in the rendering process, were captured from real-world hand-object manipulation using a marker board attached to the manipulated object.

**HO-3D.** We extend our analysis on real-world data by considering the public HO-3D [10] dataset, comprising real images of hand-held YCB objects that are involved in challenging rotational motions. Furthermore, this dataset presents more variability in terms of background, object occlusion and lightning conditions. Each sequence is provided with 30 fps 640x480 RGB-D frames. This dataset offers ground-truth 6D object pose labels for benchmarking while it lacks annotations of the 6D object velocity.

We selected 18 among all the HO-3D sequences, excluding those showing YCB objects for which a DOPE weight was not available and those containing discontinuities in the pose trajectory, hence not suitable for tracking purposes. We refer to our public software implementation[2] for more details.

For further qualitative evaluation, we provide real unlabeled sequences acquired with an Intel RealSense D415 camera, representing a scenario similar to that of Fast-YCB.

### B. Implementation details

We assume that the RGB-D input stream is available at 30 fps. To obtain flow frames $\{F_k\}$ we use the readily available NVIDIA Optical Flow SDK which provides real-time performance with small CUDA cores utilization [26]. Since we consider a scenario where the output of the segmentation and pose estimation networks is only available at low frame rate, we run Mask R-CNN and DOPE at 5 fps. This configuration corresponds to a delay of 200 ms or, equivalently, to a value of $N_s = N_p = 6$. The actual values of the covariance matrices $Q_*$ and $R_*$ were chosen empirically in order to ensure a fast response with minimal overshoot, residual oscillations and noise in the output. They can be found in our public software implementation[2]. The same set of parameters has been used in all the experiments independently of the considered dataset or object.

---

[1] https://github.com/hsp-iit/fast-ycb
[2] https://github.com/hsp-iit/roft

TABLE I: Results on the Fast-YCB dataset: ADD-AUC and RMSE Cartesian and angular errors for several methods.

| metric | ADD-AUC (%) | | | | RMSE $e_t$ (cm) | | | | RMSE $e_a$ (deg) | | | |
|---|---|---|---|---|---|---|---|---|---|---|---|---|
| method | DOPE [23] | ROFT (ours) | PoseRBPF [6] | se(3)-TrackNet [7] | DOPE [23] | ROFT (ours) | PoseRBPF [6] | se(3)-TrackNet [7] | DOPE [23] | ROFT (ours) | PoseRBPF [6] | se(3)-TrackNet [7] |
| 003_cracker_box | 54.92 | **78.50** | 68.94 | 63.02 | 5.1 | **2.5** | 2.6 | 7.9 | 28.325 | **7.545** | 38.455 | 31.729 |
| 004_sugar_box | 60.01 | 81.15 | **82.78** | 73.70 | 5.1 | 2.4 | **1.9** | 4.4 | 34.636 | **8.391** | 17.073 | 31.623 |
| 005_tomato_soup_can | 64.14 | 79.00 | 75.93 | **80.82** | 4.7 | 2.9 | **1.1** | 3.3 | 29.556 | **16.571** | 63.811 | 26.282 |
| 006_mustard_bottle | 57.20 | 73.10 | **82.92** | 74.83 | 8.6 | 3.1 | **2.0** | 11.9 | 36.132 | **13.292** | 18.418 | 35.480 |
| 009_gelatin_box | 60.01 | **74.26** | 11.32 | 69.00 | 34.4 | **4.7** | 13.3 | 35.5 | 27.936 | **16.368** | 65.857 | 32.311 |
| 010_potted_meat_can | 57.03 | 73.87 | **87.29** | 71.30 | 6.1 | 3.1 | **1.2** | 19.2 | 31.444 | **10.838** | 23.758 | 44.712 |
| ALL | 58.83 | **76.59** | 68.10 | 72.06 | 15.1 | **3.2** | 5.7 | 17.6 | 31.491 | **12.675** | 42.979 | 34.155 |

We performed our experiments on a mobile computer equipped with an Intel Core i9-10980HK CPU and an NVIDIA RTX 2080S GPU.

### C. Comparison with the State of the Art

We compare our method against two state-of-the-art object pose trackers from RGB-D images: PoseRBPF and se(3)-TrackNet. Although the authors suggest that se(3)-TrackNet can be executed without periodic re-initialization, we experimentally observed that it looses track on both the considered datasets. For this reason, and in order to have a fair comparison with our method, we exploit DOPE predictions, delayed and at low frame rate, to re-initialize it periodically. We remark that the re-initialization stage does not affect the real-time performance. Differently from se(3)-TrackNet, PoseRBPF cannot run in real-time and it is reported to achieve the highest frame rate of 17.5 fps in the configuration with 50 particles [6]. In our setup, this configuration could not achieve more than 13.5 fps, which implies to execute a filtering step every 2.2 frames. As this is not feasible due to the discrete nature of a dataset sampled at 30 fps, we executed a filtering step every 2 frames. The resulting configuration corresponds to an equivalent frame rate of 15 fps. Similarly to se(3)-TrackNet, we observed that PoseRBPF can lose the track. However, the provided mechanism for re-initialization requires more than 2 seconds on average, due to the need for exploring the state space, making it incompatible with the claimed frame rate. For this reason we could not perform the re-initialization in our experiments.

We initialize all the methods using the first available pose from the DOPE pose estimation network.

## V. EXPERIMENTAL RESULTS

In this section we perform a quantitative evaluation of the performance of the proposed method on the Fast-YCB and HO-3D datasets. The analysis includes the standard ADD-AUC metric[1], as well as the pose RMSE (root mean square) [17] tracking error. The latter is more susceptible to spikes in the output trajectory, and we deem important to consider it for robotics applications where perception is used to close the control loop. Velocity RMSE tracking errors are also discussed for the Fast-YCB dataset for which ground-truth velocities are available. Furthermore, we provide an ablation study where we selectively disable key components of our pipeline, and considerations on computation times. Qualitative results are also provided in Fig. 3.

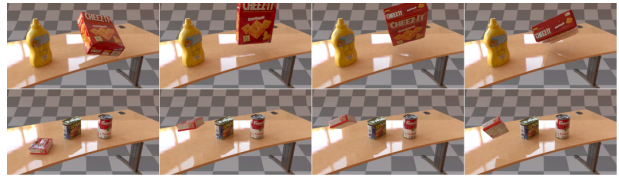

Fig. 2: Sample frames taken from the Fast-YCB dataset. The time length of the reported frames is approximately 1 second.

### A. Results on Fast-YCB

*1) Performance of 6D object pose tracking:* In Table I, we compare the performance of our pipeline against the state of the art in 6D object pose tracking from RGB-D images. We also compare against the pose estimation network DOPE (also our pose measurements input) running at 5 fps as baseline (we compare the ground-truth signal at each $k$ with the last available output from the network). We remark that the RMSE pose error is split in two terms, the Cartesian error $e_t$ and the angular error $e_a$ evaluated as the geodesic [27] between the ground-truth and the estimated orientation in $[0, \pi)$. We do not consider the the ADI-AUC metric [1] for symmetric objects given that all the considered objects are not symmetric.

Considering all the objects on average, our method achieves the best performance according to all the metrics and gets an average Cartesian error in the order of few centimeters and an average angular error of approximately 13 degrees. Remarkably, our method outperforms the others in terms of angular error not only on average but when considering each object separately. The angular error is reduced by approximately 71%, 63% and 60% on average with respect to PoseRBPF, se(3)-TrackNet and DOPE respectively.

We recall that our experiments with se(3)-TrackNet involve re-initialization from DOPE predictions once they are available (at 5 fps). Although we could have run the algorithm while avoiding re-initialization, we found that this configuration leads to early loss of tracking, with an average ADD-AUC percentage of 50.55% (i.e. 21.5 points less than the performance achievable with re-initialization).

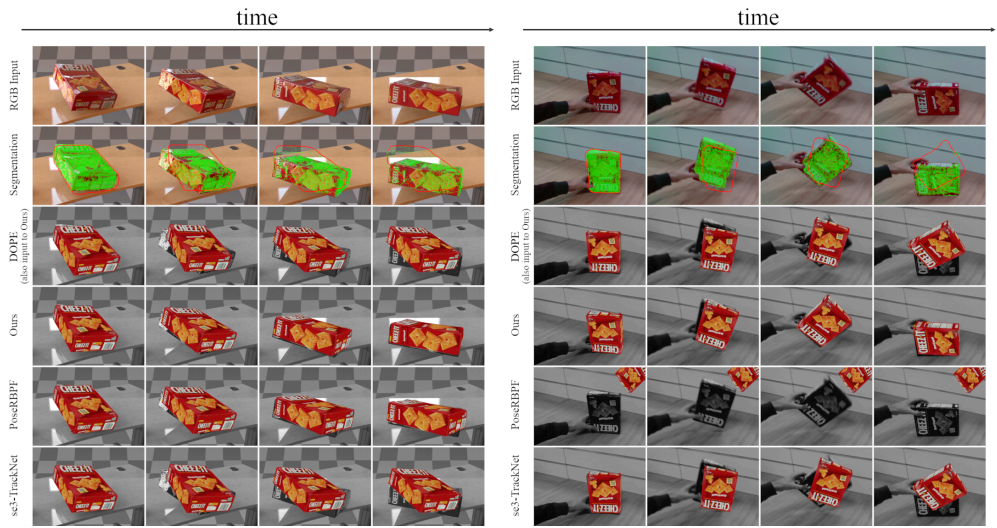

Fig. 3: Qualitative results for the object "003_cracker_box". On the left: frames from Fast-YCB. On the right: frames acquired with a RealSense D415 camera. The time length of the object motion is ≈ 1 second. The red contour in the Segmentation row represents the stream of delayed masks $\{M^d\}$, the green dots represent the masks $\{M\}$ propagated using the optical flow.

*2) Performance of 6D object velocity tracking:* In Table II we report the RMSE errors for the linear and angular velocities $\mathcal{V}$ estimated as described in Sec. III-B. We compare the obtained performance with and without the optical flow-aided segmentation mechanism that we presented in Sec. III-A.

In terms of absolute errors, our method gets approximately 11 cm/s and 32 deg/s errors on average. We deem these results as fairly accurate considering that the objects motion reaches velocities up to 63 cm/s and 266 deg/s. With respect to the scenario in which we use non-synchronized masks (w/o segm. sync.), when using segmentation masks refined with the optical flow, the tracking errors reduce considerably (on average, 45% and 54% less error for the linear and angular parts respectively).

*3) Qualitative results:* In Fig. 3, we provide qualitative results for the object "003_cracker_box". We compare pose predictions by superimposing the textured mesh of the object, transformed according to the estimated pose, on the input frames. Our method provides well-synchronized estimates in case of fast motions from both a Fast-YCB sequence and one acquired in a real setting using a RealSense D415 camera. The estimates from DOPE (also our pose measurements) and se(3)-TrackNet show lack of synchronization especially in terms of orientation. The performance of PoseRBPF is similar to ours on the synthetic sequence, but it looses track on the real sequence.

These results can be largely attributed to the usage of the optical flow via the object velocity $\mathcal{V}$. The velocity conveys the 6D direction of motion to the Unscented Kalman filter that is valuable in presence of fast motions. Methods based on particle filtering, as PoseRBPF, usually do not employ this kind of information and explore the state space via random walks. Conversely, end-to-end approaches for object tracking, as se(3)-TrackNet, predict the relative 6D pose between consecutive time steps directly from images. This task is rather complex to be accomplished and might result in underestimation of the actual motion or tracking loss. Our approach, instead, uses optical flow which solves the simpler task of estimating the motion of pixels in 2D space. This information is later converted to the 6D domain using a filtering based approach that does not require training.

TABLE II: Results on the Fast-YCB dataset: RMSE linear and angular velocity errors.

| metric | RMSE $e_v$ (cm/s) | | RMSE $e_\omega$ (deg/s) | |
|---|---|---|---|---|
| method | ROFT | ROFT w/o segm. sync. | ROFT | ROFT w/o segm. sync. |
| 003 | **5.237** | 8.469 | **18.250** | 38.163 |
| 004 | **8.500** | 15.996 | **24.949** | 65.321 |
| 005 | **9.592** | 21.033 | **41.212** | 78.006 |
| 006 | **7.227** | 16.409 | **26.553** | 65.494 |
| 009 | **22.277** | 32.428 | **46.374** | 94.324 |
| 010 | **6.388** | 23.194 | **26.083** | 67.406 |
| ALL | **11.409** | 20.931 | **32.119** | 70.164 |

In the second row of the same figure we represent the stream of non-synchronized masks $\{M_k^d\}$ from Mask R-CNN (red silhouette) and compare it with the output of the optical flow-aided mask synchronization mechanism as described in III-A (green pixels). It can be noticed that, although the obtained masks are discontinuous, they provide a reasonable source of pixel coordinates belonging to the object as required by the 6D velocity tracking stage described in Sec. III-B.

*4) Ablation study:* In Table IV, we analyze the effect of selectively disabling key features of our pipeline and using the ground-truth segmentation (GT segm.) and/or pose measurements (GT pose). The comparison also includes delayed DOPE predictions at 5 fps (indicated as "DOPE") and DOPE predictions on all the frames at 30 fps without delay (indicated as "DOPE (ideal)").

Excluding DOPE (ideal) and experiments with ground-truth inputs, our full pipeline reaches the best performance according to the ADD-AUC (76.59%) and RMSE angular error metrics (12.675 deg). While the pose synchronization mechanism increases the RMSE Cartesian error by only 0.2 cm, it reduces the average angular error by approximately 5 degrees. On the other hand, the outlier rejection mechanism mostly affects the Cartesian error, reducing it by approximately 7 cm. Remarkably, our pipeline achieves better RMSE errors even when compared with the DOPE (ideal). This phenomenon can be explained by the presence of the outlier rejection mechanism.

TABLE III: Results on the HO-3D dataset: ADD-AUC and RMSE Caretesian and angular errors for several methods.

| metric | ADD-AUC (%) | | | | RMSE $e_t$ (cm) | | | | RMSE $e_a$ (deg) | | | |
|---|---|---|---|---|---|---|---|---|---|---|---|---|
| method | DOPE [23] | ROFT (ours) | PoseRBPF [6] | se(3)-TrackNet [7] | DOPE [23] | ROFT (ours) | PoseRBPF [6] | se(3)-TrackNet [7] | DOPE [23] | ROFT (ours) | PoseRBPF [6] | se(3)-TrackNet [7] |
| 003_cracker_box | 48.81 | **70.52** | 54.21 | 52.07 | 9.8 | **3.3** | 3.8 | 6.8 | 45.918 | **23.085** | 67.665 | 42.743 |
| 004_sugar_box | 60.28 | **73.80** | 60.87 | 67.20 | 10.5 | 3.1 | **2.8** | 8.9 | 47.169 | **27.243** | 71.168 | 45.790 |
| 006_mustard_bottle | 36.32 | 56.29 | **59.09** | 51.89 | 17.0 | 6.6 | **2.9** | 16.3 | 78.987 | **40.107** | 86.246 | 76.446 |
| 010_potted_meat_can | 43.27 | 46.23 | **53.87** | 53.67 | 13.1 | 9.0 | **3.2** | 14.1 | 57.547 | **41.618** | 107.586 | 54.454 |
| ALL | 49.34 | **63.31** | 57.82 | 58.36 | 12.7 | 5.7 | **3.1** | 11.9 | 57.761 | **33.437** | 83.318 | 55.462 |

TABLE IV: Performance of our pipeline on the Fast-YCB dataset when key components are selectively disabled.

| variant | ADD-AUC (%) | RMSE $e_t$ (cm) | RMSE $e_a$ (deg) |
|---|---|---|---|
| ROFT w/ GT segm. and pose | 84.51 | 2.5 | 10.591 |
| ROFT w/ GT pose | 83.97 | 2.6 | 11.507 |
| ROFT w/ GT segm. | 77.08 | 3.1 | 11.901 |
| ROFT | **76.59** | 3.2 | **12.675** |
| ROFT w/o pose sync. | 74.83 | **3.0** | 17.060 |
| ROFT w/o outlier rej. | 74.79 | 10.8 | 14.248 |
| ROFT w/o segm. sync. | 64.69 | 5.4 | 27.497 |
| ROFT w/o velocity | 54.23 | 19.8 | 36.492 |
| ROFT w/o pose | 13.27 | 15.9 | 80.012 |
| DOPE [23] (ideal) | 84.94 | 8.7 | 16.277 |
| DOPE [23] | 58.83 | 15.1 | 31.491 |

TABLE V: ADD-AUC and RMSE errors averaged on all the selected HO-3D sequences when ground-truth inputs are used.

| variant | ADD-AUC (%) | RMSE $e_t$ (cm) | RMSE $e_a$ (deg) |
|---|---|---|---|
| ROFT | 63.31 | 5.7 | 33.437 |
| ROFT w/ GT segm. | 62.54 | 5.5 | 34.693 |
| ROFT w/ GT pose | 82.00 | 2.3 | 15.820 |
| ROFT w/ GT segm. and pose | 81.86 | 2.1 | 16.238 |

If velocity measurements $\{\mathcal{V}_k\}$ are not used, the performance degrades considerably because the filtering process performs corrections based on a new pose every $N_p$ steps only. Also, using DOPE alone at 5 fps would give better results in this scenario. This result confirms the importance of using the optical flow via the object velocity $\mathcal{V}$. Conversely, when velocities are used without pose measurements $\{T_k^d\}$, the performance of our method is the lowest due to the noise in the velocities, which gets integrated over time.

Overall, these results show that each component of our pipeline contributes to the overall performance. Remarkably, only a combination of pose and velocity measurements justifies the necessity to use a Kalman filtering approach for information fusion.

Experiments with ground-truth inputs show that the performance of our pipeline can be further increased. This suggests that pipelines with a certain degree of modularity, as the one proposed, where each component plays a clear and explainable role, can be more conveniently improved by possibly replacing some of their components with better ones.

### B. Results on HO-3D

In Table III we compare the performance of our method against the baseline (DOPE) and the state of the art on the HO-3D dataset. Given the considerable amount of occlusions due to the interaction between objects and the human hand, this dataset poses more challenges to DNNs for object segmentation, pose estimation and tracking. As a result, the quality of the PoseRBPF and se(3)-TrackNet predictions and of the inputs to our pipeline reduces considerably.

According to the ADD-AUC metric, even the best performance, reached by our pipeline, stops at approximately 64%. The best performance in terms of RMSE Cartesian error is instead obtained by PoseRBPF. On the other hand, our pipeline still achieves the best performance in terms of RMSE angular error for each object, with average reductions in the error by approximately 60%, 44% and 40% compared to PoseRBPF, DOPE and se(3)-TrackNet respectively.

Notably, as shown in Table V, the performance of our method improves by increasing the quality of the inputs. E.g., using ground-truth poses leads to an increase of more than 18 points of the ADD-AUC metric and a reduction of both the RMSE errors to less than half the original error.

### C. Limitations

We have observed that out-of-plane object rotations (see Fig. 4a) can sometimes result in the underestimation of the optical flow motion vectors $F_k$. The difficulty to track the pixels motion comes from object self-occlusions and causes less precise tracking of the angular velocity $\omega_k(\mathcal{V}_k)$ and, therefore, of the object pose. We provide an example of the phenomenon in Fig. 4b. Nevertheless, we observed that this condition mainly affects the magnitude of the angular velocity rather than its direction, as the final orientation of the object in Fig.4b shows. We expect that an online adaptation of the covariance matrices $R_*$ in Eq. (13) could help improve the tracking performance by giving more importance to the velocity measurements when required. We plan to investigate the design and the validation of such adaptation scheme in future research.

The proposed tracking framework has been designed to track the relative pose and the relative velocity between the camera reference frame and the reference frame attached to the object of interest. This choice implies that it can also be used in the presence of a moving camera. Nevertheless, our experiments have been conducted on the Fast-YCB and HO-3D datasets, which have been acquired using a stationary camera and do not include camera ego-motion. We plan to further investigate

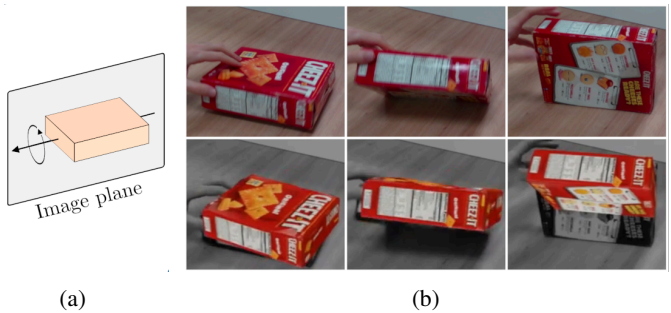

Fig. 4: Example of out-of-plane object rotation. (b) Top: input images. Bottom: example of imprecise estimates in presence of out-of-plane object rotations.

the effect of ego-motion on the proposed pipeline in future research.

*D. Computation time*

In our setup, ROFT achieves 96.2 fps compared to 142.4 fps obtained by se(3)-TrackNet and 13.5 fps achieved by PoseRBPF (with 50 particles). The frame rate has been evaluated as the inverse of the mean iteration time averaged on all the sequences of both Fast-YCB and HO-3D datasets. All the methods have been executed on the same hardware. Our method and se(3)-TrackNet exceed the real-time performance requirement (30 fps).

## VI. CONCLUSION

In this work, we proposed a Kalman filtering-based approach (ROFT) for 6D object pose and velocity tracking of fast-moving objects from RGB-D sequences. ROFT exploits real-time optical flow to synchronize, track and filter predictions from low frame-rate deep networks for instance segmentation and 6D pose estimation running in a scenario with limited computational resources. We introduced a novel dataset (Fast-YCB) comprising fast object motions, which provides ground-truth 6D object pose and velocity annotations. Experiments on the Fast-YCB and HO-3D datasets showed that our approach allows improving the pose tracking performance, when compared against the state-of-the-art object pose tracking algorithms (se(3)-TrackNet and PoseRBPF).